\documentclass[letterpaper, 10 pt, conference]{ieeeconf}

\usepackage{diagbox}
\usepackage{enumerate}
\usepackage{cite}
\usepackage{graphicx}
\usepackage{lmodern}
\usepackage{anyfontsize}
\usepackage{hyperref}
\usepackage{CJK}
\usepackage{url}

\IEEEoverridecommandlockouts                              

\title{\LARGE \bf
	Comparison and Evaluation of 2D and 3D Range Sensors
}

\author{Qingwen Xu and S\"oren Schwertfeger$^1$
	\thanks{$^1$Both authors are with the School of Information Science and Technology, 
		ShanghaiTech University, 200031 Shanghai, China, \hspace{2cm}
		{\tt\small [xuqw, soerensch]@shanghaitech.edu.cn}}%
}

\begin{document}

	\maketitle
	\thispagestyle{empty}
	\pagestyle{empty}

	\begin{abstract}
		For mobile robots range sensors are important to perceive the environment. Sensors that can measure in a 3D volume are especially significant for outdoor robotics, because this environment is often highly unstructured. The quality of the data gathered by those sensors influences all algorithms relying on it. In this paper thus the precision of several 2D and 2.5D sensors is measured at different ranges and different incidence angles. The results of all tests are presented and analyzed. 
	\end{abstract}
	
	
	\section*{Translation Notice}
	This is an English version of this Chinese Journal paper by the same authors:
	\begin{CJK}{UTF8}{gbsn}
		徐晴雯 and S{\"o}ren Schwertfeger. 多种激光传感器实验对比与评估. In {\it 电子设计工程}, 2018. (Qingwen Xu and S\"oren Schwertfeger. Comparison and evaluation of multiple laser sensors. In Electronic Design Engineering). \url{http://mag.ieechina.com/oa/DArticle.aspx?type=view&id=201702050}
	\end{CJK}
	
	%
	\IEEEpeerreviewmaketitle

	\section{Introduction}
	
	One essential capability of mobile robots is to perceive their environment using sensors. A very popular class for robots are range sensors, because they provide metric information about the distance to obstacles. The robots use this range information to navigate around obstacles, to localize themselves, to perceive the environment, and to build a map, especially using simultaneous localization and mapping (SLAM) techniques
	\cite{FRJ09-SI-3Dmap-DisasterCity}. 
	In this paper we evaluate the precision of several state of the art range sensors. First we concentrate on 2D laser range finders (LRF), which use a spinning mirror to reflect a laser measurement beam in different angles on a plane. The data collected with those 2D LRF sensors is often sufficient for robot to navigate in structured environments such as offices or homes. But research in outdoor robotics, for example also in the area of Safety, Security and Rescue Robotics (SSRR), has to deal with highly unstructured terrain for which we need information about distances to obstacles not just on a 2D plane but in the 3D volume surrounding the robot \cite{SSRR09-RREE-3Ddata}.
	
	Sensors providing such data in the 3D volume are often called 2.5D sensors, because they return only at most one distance (depth) per scan direction. This paper evaluates three fundamentally different types of 2.5D range sensors: structured-light RGBD sensors (Asus xTion, also called Kinect because that was the first famous brand of such type of sensors), sensors with multiple laser beams scanning in different directions at the same time (Velodyne VLP-16 and HDL-32E) as well as, mainly for comparison and ground truth, a sensor with two actuated angles that uses a single beam to slowly but very densely and accurately scan the environment (Faro Focus 3D). We also compare the results with data from two models of 2D LRF (Hokuyo sensors)
	\cite{Park2006RangeSensor}.
	
	Previous work has evaluated different sensors 
	\cite{3DSensorSurvey-Rescue-RCup10} 
	using different evaluation criteria such as accuracy
	\cite{stoyanov2011comparative}. 
	Especially time of flight cameras
	\cite{chiabrando2009sensors} 
	and structured light cameras have been reviewed a lot
	\cite{khoshelham2012accuracy} 
	\cite{rafibakhsh2012analysis}, 
	since their errors are typically considerably larger than laser scanning devices. The main disadvantage of those two range sensor approaches is that their performance quickly deteriorates the more sun light is present in the scene.
	
	There is also quite some previous work on benchmarking 3D laser range finders
	\cite{tucker2002testing} 
	\cite{ForwardModelFusion-IROS07} 
	\cite{okubo2009characterization} 
	\cite{kaartinen2012benchmarking}. 

	The paper is structured as follows: Section \ref{sec:methodology} introduces the methodology used for the sensor benchmarking, including the RANSAC algorithm and the mean squared error (MSE). Section \ref{sec:experiments} then describes the experiments preformed and provides some details regarding the implementation. The results and some analysis are presented in Section \ref{sec:results}. Section \ref{sec:conclusion} concludes this paper.

	\section{Methodology for Precision Evaluation}
	\label{sec:methodology}
	It is very important to form a criteria to objectively compare the precision of different sensors, which should be a simple and meaningful way to explain the results of the experiments. Precision is a significant feature to represent measurement resolution and has a strong relationship with reproducibility of measurement results\cite{hattori2008precision}. The main goal of this work is to compare the precision among different sensors and find if different models of the same product have similar performance.
	
	It would also have been nice to measure the accuracy of the sensors, but this is quite hard given the low deviation from the true distance that most of the sensors feature and thus left for future work.

	The precision measurement of a sensor is to how close measurements of the same feature are to each other. In the following experiments we present the sensor with a planar object and then evaluate how well the sensor data represents a plane or a line using several spatially and/ or temporally neighboring measurements from the planar object. We do this with different incidence angles between the sensor and the planar object to benchmark the influence of the incidence angle to the precision of the data. We are also repeating the experiments for different distances between the object and the sensor. Fitting the data onto lines or planes is a common way to benchmark sensors 
	\cite{Wang2001} 
	\cite{khoshelham2012accuracy} 
	\cite{PlaneUncertainty-JISR09}, 
	which is of course also related to the actual process of extracting planes in such range data 
	\cite{Pathak2009}. 
	
	The model of the object should be constructed first and then the distance between the points and the model is calculated. Model fitting can be implemented with several algorithms and the coincidence of the model and the actual object can be ignored because the emphasis of this work is precision not accuracy.
	
	In general, the main methodology can be divided into two parts. The first part is the algorithm for model fitting. We choose random sample consensus (RANSAC) because it is very robust and quite appropriate for this problem. The second part is about the evaluation of the precision, where mean squared error (MSE) is a good choice.
	
	\subsection{Random Sample Consensus Algorithm}
	
	RANSAC is an iterative method based on statistics, which is widely used in computer science. The algorithm can be easily described as follows:
	\vspace{2ex}
	
	\begin{tabular}{l}
		\hline
		RANSAC\\
		\hline
		1. Given a point set with $N$ points and fitting\\
		model (line,plane etc.) \\
		2. \textbf{Repeat} \\
		3. Choose least points that can determine the model.\\
		4. Calculate the coefficients of the model.\\
		5. Count the number of inliers (points with the \\
		distance to model less than threshold).\\ 
		6. \textbf{Until} Maximize the number of inliers.\\
		\hline
	\end{tabular}
	\vspace{2ex}
	
\begin{figure}[tb]
		\centering
		\includegraphics[width=1\linewidth]{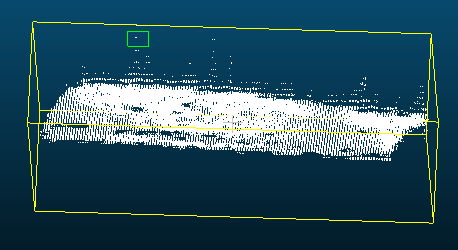}
		\caption{Point cloud data representing a plane with several outliers (e.g. point in green box).}
		\label{fig:ransac}
	\end{figure}

	RANSAC is a robust algorithm for fitting a model into data, especially in the case of disturbance. For example, as shown in Figure \ref{fig:ransac}, the point in green box is an outlier to the plane in the yellow box and there are several other outliers like this point. The least squares algorithm can not fit the accurate plane under the interference of outliers. However, the RANSAC algorithm it advantageous by being able to fit the plane without disturbance by these outliers. 
	
	In this paper, RANSAC is used to implement both line and plane fitting. Line fitting is suitable for all sensors but when it comes to dense 2.5D data, the line cut by hand will induce some error. In order to reduce the error, the width of the line cut by hand is less than double distance between adjacent points.
	
	\subsection{Mean Squared Error}
	The mean squared error (MSE) is widely used in many fields, such as signal processing, statistics and computer vision\cite{wang2009mean}. MSE is a convention and many algorithms take it as a criteria to compare their results. 
	
	Suppose that there are several sample points $S=\{(x_i,y_i,z_i)|i=1,2,...,N\}$ and the fitted plane $Ax+By+Cz+D=0$, the MSE can be described as Eq.\ref{1} and Eq.\ref{2}, which can reflect the error of the point cloud.
	\begin{equation}\label{1}
	MSE=\frac{1}{N}\sum_{i=1}^{N}d_i,
	\end{equation}
	\begin{equation}\label{2}
	d_i=\frac{Ax_i+By_i+Cz_i+D}{\sqrt{A^2+B^2+C^2}}
	\end{equation}
	
	In this experiment, MSE is chosen as the principle of precision evaluation because it features the following advantages:
	\begin{itemize}
		\item Simple. It is not trapped in parameters and uncomplicated to compute. Besides, it saves memory because square error can be computed one by one.
		\item It has a clear physical meaning. In the field of signal processing, it usually represents the energy of the error\cite{wang2009mean}. In this work, it can describe the error between points and the fitted model in a good way.
		\item It is beneficial for optimization because of its properties, such as convexity, differentiability and symmetry. 
	\end{itemize}

	\section{Experiments}
	\label{sec:experiments}
\begin{figure}[tb]
		\centering
		\includegraphics[width=1\linewidth]{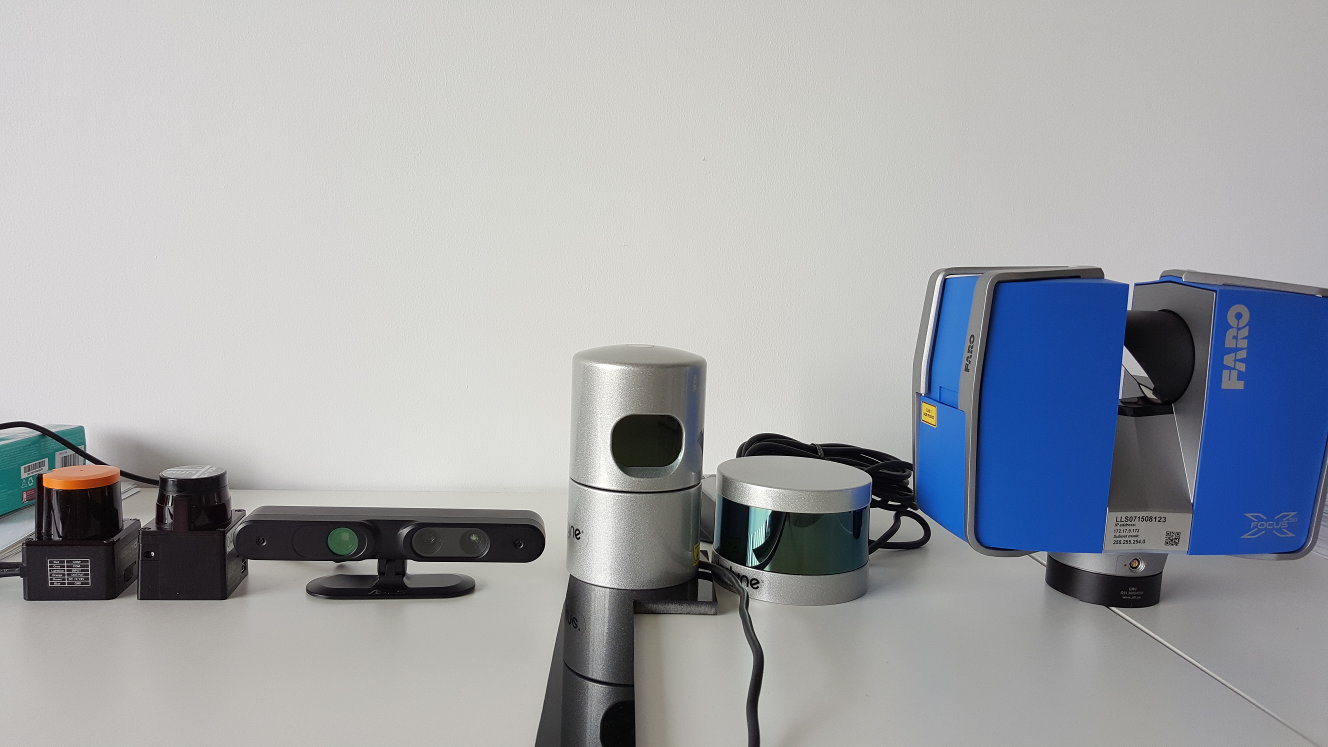}
		\caption{The sensors evaluated in this paper. They are, from left to right, Hokuyo UST-10LX, Hokuyo URG-04LX-UG01, Asus xTion, Velodyne HDL-32E, Velodyne VLP-16 and FARO Focus 3D X330.}
		\label{fig:sensors}
	\end{figure}

	Sensors for the experiment include four 2D laser scanners (two Hokuyo URG-04LX-UG01 and two Hokuyo UST-10LX) and six 2.5D sensors (two Asus xTion live, one Velodyne VLP-16, two Velodyne HDL-32E and a FARO Focus X330) as shown in Figure \ref{fig:sensors}. Some of their properties (as listed in the data-sheets) are presented in Table \ref{tab:sensors}. Since the Faro cannot be used on-line we omitted its power and connection specifications. It should be mentioned that sensors of the same type are tested to verify that there are few differences among them. According to the range of each sensor, the experiment is divided to three tests. The distances between the sensors and object are two meters, four meters and ten meters, respectively. In addition, all tests were taken in a dark environment to ensure the performance of sensors. 

	\begin{table*}[h]
		\newcommand{\tabincell}[2]{\begin{tabular}{@{}#1@{}}#2\end{tabular}}
		\centering
		\caption{\label{tab:sensors} Sensor Specifications}
		\tabcolsep=0.1cm
		\scriptsize
		\begin{tabular}{|c|c|c|c|c|c|c|}\hline
			& 04LX & 10LX & xTion & VLP-16 & HDL-32E & Focus X330 \\ \hline
			max. Range (m) & 5 & 10 & 5 & 100  & 100 & 330 \\ \hline
			scan Freq. (Hz) & 10 & 40 & 30 & 10 & 10 & 3-120 min \\ \hline
			points per scan & 768 & 1080 & 307k & 30k & 70k & up to 700M\\ \hline
			2D/ 2.5D & 2D & 2D & 2.5D & 2.5D & 2.5D & 2.5D\\ \hline
			works in sunlight & yes & yes & no & yes & yes & yes \\ \hline
			approx. price (USD) & 1,200  & 1,800 & 130 & 8,000 & 30,000 & 50,000 \\ \hline
			approx. size & 50*50 & 50*50 & 457*89& $\phi$103 & $\phi$75 & 240*200 \\ 
			(W*D*Hmm)  & *70 & *70 & *127 & *72 & *144 & *100\\ \hline
			approx. weight & 160g & 130g & 220g  & 830g & $<$2kg & 5.2kg \\ \hline
			DC power  & 5V & 12V/ 24V &  5V & 12V & 12V & / \\ 
			consumption & 2.5W & 3.6W &  2.5W & 8W & 12W & / \\ \hline 
			connection type & USB & RJ45 & USB & RJ45 & RJ45 & /  \\ \hline
			
		\end{tabular}
	\end{table*}

	\subsection{Raw Data Acquisition}
\begin{figure}[tb]
		\centering
		\includegraphics[width=1\linewidth]{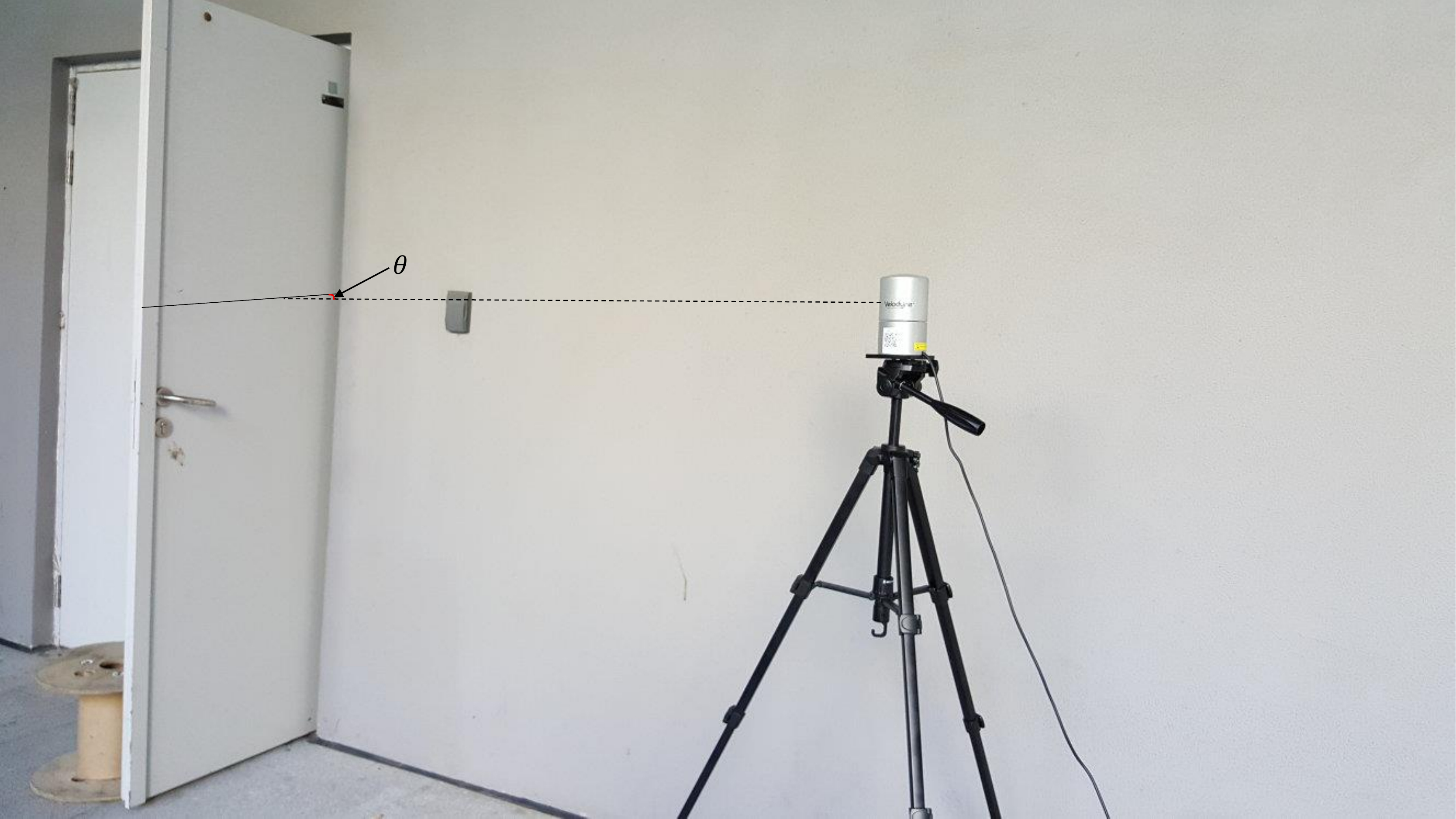}
		\caption{Set up of one experiment: Velodyne HDL-32E at 2m distance and 0$^\circ$ angle.}
		\label{fig:SetUp}
	\end{figure}

	In this work, a metal door painted with smooth lacquer is chosen as the planar target object. Figure \ref{fig:SetUp} show the set up of our experiments. Here displays the Velodyne sensor at a distance of two meters from the object. As mentioned in Section 2, the work mainly focuses on whether precision relates to different incidence angles $\theta$ between incident laser and the plane of door as shown in Figure \ref{fig:SetUp}. Each test includes three measurements at the angles $\theta=45^\circ$, $60^\circ$, $90^\circ$ of each sensor at the different distances. Rosbag is used to collect raw data taking the advantage of robot operating system (ROS)\footnote{\url{http://wiki.ros.org}}.
	\begin{table}[h]
		\centering
		\begin{tabular}{|l|}
			\hline
			\hspace*{1cm}rosbag record -O [bagname] [topic] \hspace*{1cm} \\
			\hline
		\end{tabular}
	\end{table}
	
	\subsection{Model Data}
	The type of raw data is sensor$\_$msgs/PointCloud2. Firstly, the raw data is converted to the type of PointXYZ, which can be handled by most software. PointXYZ is a triplet with coordinate $(x,y,z)$ and $z=0$ in two-dimension case. Secondly the software CloudCompare\footnote{\url{http://cloudcompare.org/}} is used to segment the model. CloudCompare is an open source and free software for 3D point cloud data processing and analysis.
	
	\begin{figure}[tb]
		\centering
		\includegraphics[width=1\linewidth]{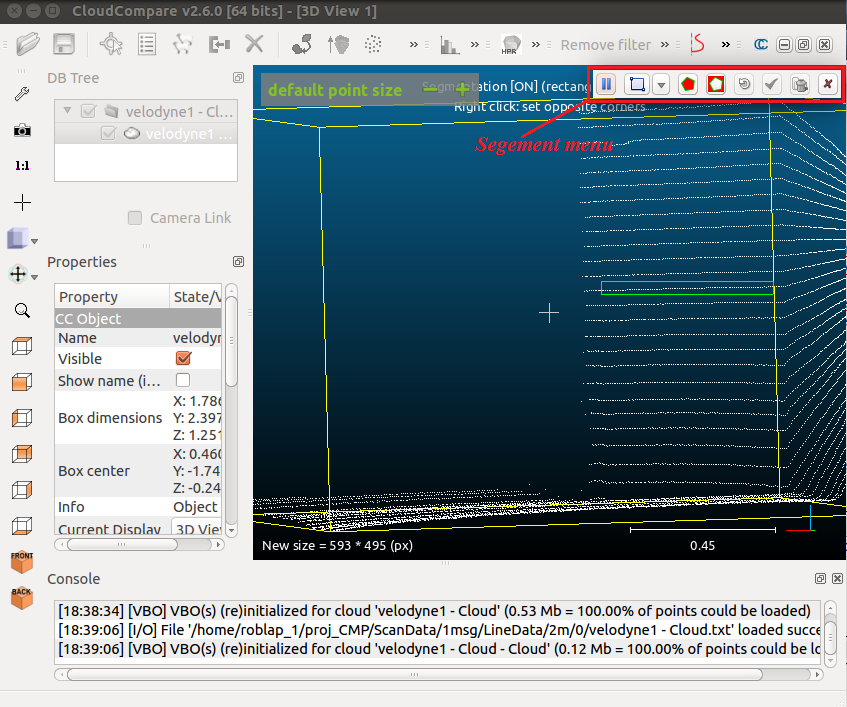}
		\caption{CloudCompare Segmentation}
		\label{fig:cloudcompare}
	\end{figure}
	
\begin{figure}[tb]
		\centering
		\includegraphics[width=0.9\linewidth]{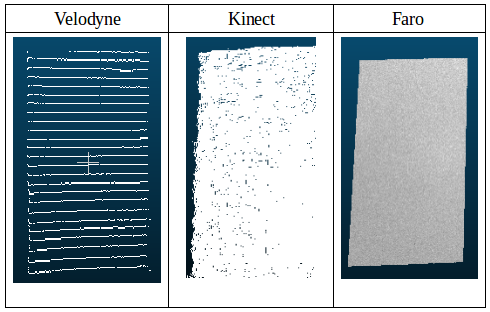}
		\caption{Data of three different 2.5D sensors.}
		\label{fig:compare}
	\end{figure}

	Figure \ref{fig:cloudcompare} shows the process of line extraction for velodyne data in Cloudcompare. There is a segment menu to cut a rectangular region in points set. The line in green box is the line that is chosen as sample data which represents the whole door data. It is obvious that cutting a line is easy from the data of the 2D laser scanner. However, the density of points influences the difficulty in segmenting a line from 2.5D data, as Figure \ref{fig:compare} shows. That is to say, a line can be easily split up from the scan of Velodyne while it is difficult to separate a line from scans of the Faro and the Kinect-like RGBD sensor. A rectangular selection with the width of 4 millimeters in CloudCompare helps to deal with the problem, which can only segment 1 to 2 lines.
	
	\subsection{Model Fitting}
	After the sensor data corresponding to the model has been extracted roughly by hand, fitting becomes much easier. The point cloud library (PCL)\footnote{\url{http://www.pointclouds.org/}} was used to implement the fitting. RANSAC is used to fit the model into the data. After the model is determined, the distance between each point and model can be calculated. 
	The MSE of all tests is calculated and  compared later.

	\section{Results and Analysis}
	\label{sec:results}
	According to the type of each sensor, the line fitting is applied to all scanners while the plane fitting is only employed for 2.5D devices (Kinect, Velodyne and Faro). According to the range of each sensor, tests with the distance of $2m$ and $4m$ are performed for all sensors while tests with the distance of $10m$ are only done for Velodyne and Faro.
	
	\subsection{Line Fitting Results}
	Nine scans of each sensor were collected and compared (three angles at three distances). The Faro tests include four different quality settings (1x being lowest and 4x highest quality). The Faro was always configured to provide the highest resolution (700 million points for a complete scan), but to only scan the relevant portion of the scene. In the following the Hokuyo URG-04LX-UG01 is abbreviated with 04LX and the Hokuyo UST-10LX is abbreviated with 10LX. If multiple sensors of the same model are tested the device number is appended to the name.
	
	\begin{table}[tb]
		\centering
		\caption{\label{l2m} Line fitting for angles at 2m}
		\begin{tabular}{|c|c|c|c|}
			\hline
			Sensor & $90^\circ$($mm^2$) & $60^\circ$($mm^2$) & $45^\circ$($mm^2$) \\
			\hline
			04LX-1 & 14.02 & 23.16 & 19.36\\
			\hline
			04LX-2 & 13.47 & 11.85 & 20.81\\
			\hline
			10LX-1 & 16.43 & 27.72 & 13.72\\
			\hline
			10LX-2 & 25.27 & 16.13 & 13.73\\
			\hline
			Kinect1 & 84.02 & 84.12 & 73.66\\
			\hline
			Kinect2 & 105.83 & 119.47 & 52.08\\
			\hline
			Velodyne16 & 45.17 & 31.26 & 22.90\\
			\hline
			Velodyne32-1 & 5.50 & 5.29 & 4.39\\
			\hline
			Velodyne32-2 & 4.62 & 5.16 & 3.55\\
			\hline
			Faro1x & 1.98 & 1.77 & 1.78\\
			\hline
			Faro2x & 1.40 & 1.56 & 1.63\\
			\hline
			Faro3x & 1.83 & 1.46 & 1.42\\
			\hline
			Faro4x & 1.32 & 1.44 & 1.47\\
			\hline
		\end{tabular}
	\end{table}
	
	Precision of each sensor among different angles does not differ a lot as seen in Table \ref{l2m}. Besides, the performance of sensors of the same type does not differ a lot. The inter-device comparison shows that the Faro has the best performance in the test, even with the lowest quality. However, one scan of Faro takes several minutes depending on quality level while one scan of other sensors only took about several milliseconds. According to the performance of sensors, the descending order is Faro, Velodyne32, Hokuyo laser scanners, Velodyne16 and Kinect. There is not obvious disparity between Hokuyo URG-04LX and UST-10LX in this test. Besides, Faro of different quality also performs in a similar way.
	
	\begin{table}[tb]
		\centering
		\caption{\label{l4m} Line fitting for angles at 4m}
		\begin{tabular}{|c|c|c|c|}
			\hline
			Sensor & $90^\circ$($mm^2$) & $60^\circ$($mm^2$) & $45^\circ$($mm^2$) \\
			\hline
			04LX-1 & 44.41 & 47.53 & 33.22\\
			\hline
			04LX-2 & 43.18 & 39.30 & 57.53\\
			\hline
			10LX-1 & 25.16 & 19.92 & 10.27\\
			\hline
			10LX-2 & 27.95 & 9.72 & 25.20\\
			\hline
			Kinect1 & 1308.43 & 473.66 & 309.70\\
			\hline
			Kinect2 & 405.25 & 928.20 & 380.34\\
			\hline
			Velodyne16 & 37.75 & 32.61 & 36.24\\
			\hline
			Velodyne32-1 & 3.83 & 8.45 & 5.04\\
			\hline
			Velodyne32-2 & 7.97 & 9.18 & 11.38\\
			\hline
			Faro1x & 1.69 & 1.77 & 3.23\\
			\hline
			Faro2x & 7.26 & 1.50 & 1.35\\
			\hline
			Faro3x & 1.55 & 3.61 & 1.31\\
			\hline
			Faro4x & 1.55 & 1.23 & 1.46\\
			\hline
		\end{tabular}
	\end{table}
	
	From Table \ref{l4m}, it is seen that difference of the performance of Hokuyo URG-04LX and UST-10LX is obvious because the maximum range of Hokuyo URG-04LX-UG01 is five meters. Hokuyo UST-10LX performs in a normal way while the precision of URG-04LX decrease a lot. The comparison between Table \ref{l2m} and Table \ref{l4m} shows that some sensors like Hokuyo UST-10LX, Velodyne16, Velodyne32 and Faro have the same performance with the distance of $2m$ and $4m$ while Kinect and Hokuyo URG-04LX perform worse in distance of $4m$ than $2m$. It should be noted that the maximum range of Kinect and Hokuyo URG-04LX are both close to five meters, which can explain the results. In addition, there is still no obvious difference between the different incidence angles because the measurement distance is quite small and much less than the range for the other sensors.
	
	\begin{table}[tb]
		\centering
		\caption{\label{l10m} Line fitting for angles at 10m}
		\begin{tabular}{|c|c|c|c|}
			\hline
			Sensor & $90^\circ$($mm^2$) & $60^\circ$($mm^2$) & $45^\circ$($mm^2$) \\
			\hline
			Velodyne16 & 32.15 & 21.36 & 4.67\\
			\hline
			Velodyne32-1 & 5.02 & 3.39 & 7.48\\
			\hline
			Velodyne32-2 & 3.00 & 4.55 & 5.37\\
			\hline
			Faro1x & 1.70 & 1.51 & 1.41\\
			\hline
			Faro2x & 0.89 & 1.40 & 1.27\\
			\hline
			Faro3x & 1.67 & 1.34 & 1.20\\
			\hline
			Faro4x & 0.90 & 1.27 & 1.24\\
			\hline
		\end{tabular}
	\end{table}
	
	From Table \ref{l10m} it is apparent that all sensors provide good performance in the test of the distance of $10m$. The maximum range of the Velodyne sensors is over 80 meters and of Faro is over 300 meters. Therefore, each sensor can perform well in the distance of $10m$. However, there is some abnormal data in Table \ref{l10m}. The most serious is the velodyne16 with angle of $45^\circ$, which is too small. When re-examining the pre-processing data, the number of points used by line fitting in this case is only $9$, which is the cause of the low mean squared error.
	
	Sensors of the range over $10m$ are all 2.5D laser scanners. It is not necessary to do line fitting in longer distance. Plane fitting with distance from $2m$ to much higher distances is used now to better compare the sensors and to verify whether the different incidence angles have an influence on the precision of the collected data.
	
	\subsection{Plane Fitting Results}
	According to Table \ref{p2m}, each sensor has different performance from that in line fitting. However, it seems that there is also not an obvious law between the angles and the mean squared error, which means that each sensor has a relatively high precision at a short distance. In this test, the MSE of the Velodyne and the Kinect are quite similar while Faro performs better. Besides, the mean square errors of the different Faro quality settings are also about the same, which attributes to the short distance.

	\begin{table}[tb]
		\centering
		\caption{\label{p2m} Plane fitting for angles at 2m}
		\begin{tabular}{|c|c|c|c|}
			\hline
			Sensor & $90^\circ$($mm^2$) & $60^\circ$($mm^2$) & $45^\circ$($mm^2$) \\
			\hline
			Kinect1 & 106.73 & 65.22 & 111.09\\
			\hline
			Kinect2 & 124.70 & 65.08 & 44.01\\
			\hline
			Velodyne16 & 58.45 & 76.35 & 68.19\\
			\hline
			Velodyne32-1 & 55.07 & 42.38 & 18.22\\
			\hline
			Velodyne32-2 & 73.36 & 80.32 & 46.51\\
			\hline
			Faro1x & 0.62 & 0.54 & 0.50\\
			\hline
			Faro2x & 0.46 & 0.35 & 0.37\\
			\hline
			Faro3x & 0.34 & 0.28 & 0.24\\
			\hline
			Faro4x & 0.40 & 0.26 & 0.24\\
			\hline
		\end{tabular}
	\end{table}
	
	In Table \ref{p4m} for the 4$m$ distance, the MSE of the Kinect again gets larger, obviously for the sake of the maximum range of the Kinect. The other sensors also have a good performance in this test because their range is much longer than 4 meters. Though the MSE of Velodyne-16e is a little larger than Velodyne-32e, both of their performances 
	can be considered as passable. In addition, the two Velodyne-32e have similar performance and the performance of Faro is the best, even at the lowest quality.
	
	\begin{table}[tb]
		\centering
		\caption{\label{p4m} Plane fitting for angles at 4m}
		\begin{tabular}{|c|c|c|c|}
			\hline
			Sensor & $90^\circ$($mm^2$) & $60^\circ$($mm^2$) & $45^\circ$($mm^2$) \\
			\hline
			Kinect1 & 745.13 & 914.40 & 475.07\\
			\hline
			Kinect2 & 690.22 & 441.29 & 498.63\\
			\hline
			Velodyne16 & 90.25 & 52.39 & 74.12\\
			\hline
			Velodyne32-1 & 40.54 & 32.02 & 24.61\\
			\hline
			Velodyne32-2 & 53.20 & 27.56 & 34.50\\
			\hline
			Faro1x & 0.61 & 0.55 & 0.48 \\
			\hline
			Faro2x & 0.45 & 0.45 & 0.40 \\
			\hline 
			Faro3x & 0.39 & 0.38 & 0.29 \\
			\hline
			Faro4x & 0.42 & 0.32 & 0.31 \\
			\hline
		\end{tabular}
	\end{table}
	
	Table \ref{p10m} shows the data for the $10m$ tests. It seems that there is little difference of MSE between the test at four meters and at ten meters. However, the number of points for fitting the plane is much less in ten meters. As we all know, three non collinear points determine a plane. If more points in this plane are provided, fitting would be more accurate. So the performance of these sensors is better in a nearer distance. 
	
	\begin{table}[tb]
		\centering
		\caption{\label{p10m} Plane fitting for angles at 10m}
		\begin{tabular}{|c|c|c|c|}
			\hline
			Sensor & $90^\circ$($mm^2$) & $60^\circ$($mm^2$) & $45^\circ$($mm^2$) \\
			\hline
			Velodyne16 & 60.40 & 64.66 & 36.98\\
			\hline
			Velodyne32-1 & 43.86 & 40.23 & 46.91\\
			\hline
			Velodyne32-2 & 56.40 & 97.10 & 50.20\\
			\hline
			Faro1x & 0.68 & 0.57 & 0.51 \\
			\hline
			Faro2x & 0.51 & 0.44 & 0.38 \\
			\hline 
			Faro3x & 0.46 & 0.38 & 0.34 \\
			\hline
			Faro4x & 0.38 & 0.31 & 0.34 \\
			\hline
		\end{tabular}
	\end{table}
	
	Table \ref{p71m} shows another test only about Faro scanner. Even though the Velodyne sensors can reach this distance, their data is too sparse in the vertical direction to achieve plane fitting at that range. During this test, the object is an acrylic plate with protective-paper on its surface. The resolution of Faro was set to maximum and the quality was set to four levels like the above tests. A region about $0.4m\times 0.4m$ was extracted as samples to implement the plane fitting and MSE calculation. In this case, it is obviously that the MSE is smaller with the ascent of the quality of Faro scanner. In addition, we can find that MSE is bigger when the angle is not $0^\circ$, especially in the angle $60^\circ$. Another noteworthy point is that scans with different quality take different time. Suppose the time of scan with lowest quality is 1, then the time of scan with $2x,3x,4x$ is about $1.78,3.32,6.41$ respectively.
	
	\begin{table}[tb]
		\centering
		\caption{\label{p71m} Plane fitting for angles at 71m}
		\begin{tabular}{|c|c|c|c|}
			\hline
			Sensor & $90^\circ$($mm^2$) & $45^\circ$($mm^2$) & $30^\circ$($mm^2$) \\
			\hline
			Faro1x & 2.00 & 4.38 & 6.44 \\
			\hline
			Faro2x & 1.50 & 2.35 & 7.27 \\
			\hline 
			Faro3x & 0.47 & 2.33 & 7.62 \\
			\hline
			Faro4x & 0.23 & 0.73 & 5.69 \\
			\hline
		\end{tabular}
	\end{table}
	
	\section{Conclusion}
	\label{sec:conclusion}
	This work uses the RANSAC algorithm to fit line and plane models to point clouds gathered by different sensors. Then the MSE between each point and model was computed to represent the precision of each scan. The precision of each sensor deteriorates with greater distances, especially when the ranges come close to the maximum range of their respective sensor.
	
	In general, the experiments show that there is not a direct relation between precision and the incidence angles when the distance is no more than $10m$, because each sensor has a good performance within its ranges. However, in the experiment of Faro with distance $71m$, it is obvious that angles have some influence on the MSE. Though it seems that there is little difference between angle $45^\circ$ and $60^\circ$, MSE in these two angles are larger than that in angle $0^\circ$. Therefore, it can be concluded that precision of Faro is lower when there is a nonzero angle between the object and the scanner in long distance measurement.
	
	Besides, different sensors have different precision in their ranges. The results shows that the Faro scanner has the highest precision, followed by the Velodyne, Hokuyo and Kinect. However, the scan time of Faro is the longest and it cannot be used as real-time sensor. Therefore, it depends on the practical situation to choose a suitable laser scanner.

	\bibliographystyle{IEEEtran.bst}
	
	\bibliography{references,References,References_old} 

\end{document}